\documentclass[10pt,twocolumn,letterpaper]{article}

\usepackage[pagenumbers]{cvpr} 

\usepackage{cuted}
\usepackage{capt-of}
\usepackage{multirow}
\usepackage{bm}

\newcommand{\fref}[1]{Figure \ref{#1}}

\newcommand{\sref}[1]{Section \ref{#1}}  

\definecolor{cvprblue}{rgb}{0.21,0.49,0.74}
\usepackage[pagebackref,breaklinks,colorlinks,allcolors=cvprblue]{hyperref}

\def\paperID{368} 
\def\confName{3DV\xspace}
\def\confYear{2027\xspace}

\title{Hand-4DGS: Feed-Forward 3D Gaussian Splatting for 4D Hand Reconstruction from Egocentric Videos}

\author{
\mbox{Jeongmin Bae$^{1\dagger}$ \quad
Seoha Kim$^{2\dagger}$ \quad
Marc Pollefeys$^{3,4}$ \quad
Mahdi Rad$^{4}$}\\
\mbox{Youngjung Uh$^{1\ddagger}$ \quad
Taein Kwon$^{5\ddagger}$}\\[6pt]
\mbox{$^{1}$Yonsei University \quad
$^{2}$ETRI \quad
$^{3}$ETH Zurich \quad
$^{4}$Microsoft Spatial AI Lab}\\
\mbox{$^{5}$VGG, University of Oxford}\\[5pt]
\mbox{$^{\dagger}$Equal contribution.
\qquad
$^{\ddagger}$Corresponding authors.}
}

\begin{document}

\maketitle

\begin{strip}
\centering
\begin{minipage}{\textwidth}
\centering

\includegraphics[width=0.9\linewidth]{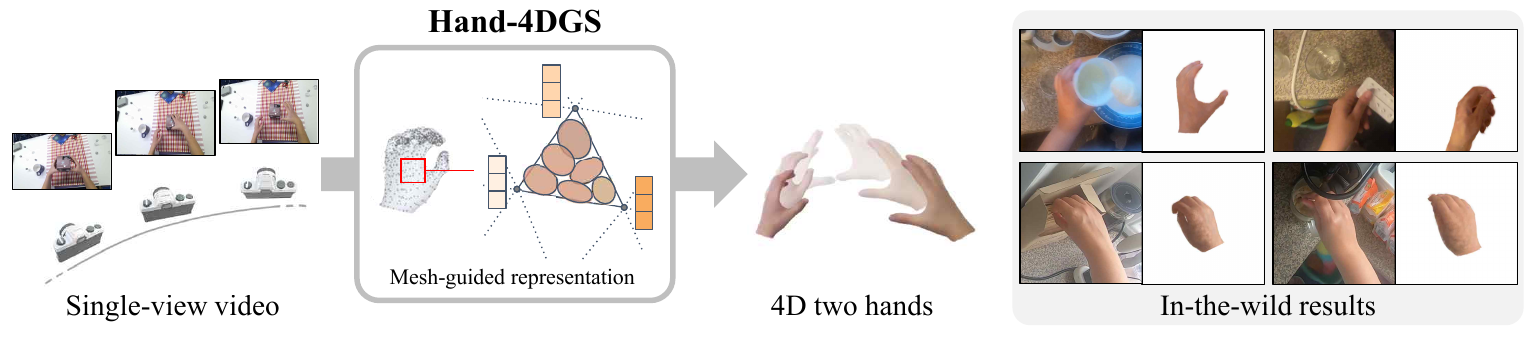}

\makeatletter
\def\@captype{figure}
\makeatother
\caption{\textbf{Overview of Hand-4DGS.}
Our feed-forward framework reconstructs dynamic 4D hands from challenging
single-view egocentric videos, without the heavy optimization time typically required by Gaussian splatting methods.
By introducing a mesh-guided representation with per-vertex embeddings, Hand-4DGS captures hand appearance and complex motion.
Our model predicts hand pose from RGB observations and adapts to unseen
videos without relying on ground-truth 3D hand pose annotations during either training or inference.}
\label{fig:teaser}
\end{minipage}
\end{strip}
\vspace{-6mm}

\begin{abstract}
Dynamic 3D hand reconstruction from egocentric videos is essential for next-generation computing platforms such as AR/VR and AI glasses. 
Despite its importance, most prior works focus either on multi-view 3D hand reconstruction or on 4D human body reconstruction. 
Egocentric 4D hand reconstruction remains difficult due to rapid hand and camera motion, hand-object and inter-hand interactions, and inherent ambiguity from single-view observations. 
To address these challenges, we introduce Hand-4DGS, a feed-forward framework for dynamic 4D hand reconstruction from egocentric videos.
Our approach incorporates a mesh-guided representation for structural priors and temporal convolutions to model dynamic motion. 
We evaluate our framework on H2O and ARCTIC, two egocentric hand-object interaction datasets, and show improvements over baselines.
Our model can efficiently adapt to unseen videos from datasets that are not included in training.
In addition, image supervision through differentiable Gaussian rasterization provides an additional signal for appearance optimization and pose adjustment during training and test-time optimization, without ground-truth 3D hand pose annotations.
\end{abstract}
\section{Introduction}
\label{sec:intro}

4D hand reconstruction is a foundational technology for the next generation of computing environments, such as augmented and virtual reality (AR/VR), AI glasses, robot manipulation, telepresence, and teleoperation, as hands provide a key cue for understanding physical interactions.
In particular, egocentric 4D hand reconstruction is gaining increasing attention as it eliminates the need for bulky sensors or multi-camera setups.

Recent studies using 3D Gaussian splatting (3DGS)~\cite{kerbl3Dgaussians} have shown promising results for reconstructing 3D hands from multiview images~\cite{guo2023handnerf, pokhariya2023manus, sun2025jghandjointdrivenanimatablehand, dong2025handsplatembeddingdrivengaussiansplatting, gaussianhand, on2025bigsbimanualcategoryagnosticinteraction}. 
However, these methods require multiple viewpoints, often ranging from 20~\cite{gaussianhand} to more than 120 views~\cite{mundra2023livehand}, to resolve the depth ambiguity inherent in a single viewpoint. 
This requirement limits their applicability in wearable scenarios, such as AR, VR, and AI glasses. 

Unlike these multi-view approaches, some recent works, HUGS~\cite{kocabas2024hugs} and EVA~\cite{hu2024expressive}, rely on off-the-shelf \textit{monocular} body pose estimators~\cite{goel2023humans, pavlakos2024reconstructing} to obtain per-frame body meshes, which serve as geometric priors to initialize the 3D Gaussians.
While effective for human body reconstruction, these approaches are less suited for 4D hand reconstruction for the following reasons. 
Body reconstruction methods often assume a static camera observing a single person, whereas egocentric scenarios involve a moving camera, severe self-occlusions, limited viewpoints, rapid hand motion, and interactions between two hands and objects.
As a result, \textit{hand} estimators often produce less reliable mesh initializations in egocentric settings, and these errors propagate to the Gaussian representation, leading to unstable or degraded reconstructions. 
Furthermore, these approaches rely on per-video optimization, which hinders real-time in-the-wild applications due to its substantial computation time.

Feed-forward 3D Gaussian approaches offer an alternative paradigm by predicting Gaussians directly from images using neural networks, enabling fast inference and generalization~\cite{charatan23pixelsplat, chen2024mvsplat, szymanowicz2024flash3d}.
However, existing methods~\cite{kwon2024ghg,GUAVA} are primarily designed for human body reconstruction rather than hand reconstruction, and they typically predict static 3D avatars from a single image or sparse-view images.
As a result, they are not suitable for modeling the dynamics required for 4D reconstruction. 
Moreover, these approaches often rely on pose estimators during both training and inference, making their performance sensitive to pose estimation errors.
Consequently, current feed-forward Gaussian methods are not well suited to fine-grained hand dynamics and remain sensitive to pose estimation quality.
To the best of our knowledge, no prior work has explored egocentric 4D hand reconstruction using a feed-forward 3D Gaussian-based approach.

To address these limitations, we present Hand-4DGS, a feed-forward 3D Gaussian framework for 4D hand reconstruction from egocentric videos. 
Our approach reconstructs hand appearance and geometry from RGB videos and supports efficient adaptation to unseen video sequences through a short test-time optimization (see \cref{fig:teaser}).
Specifically, we introduce a mesh-guided representation in which the predicted hand mesh provides a geometric prior for Gaussian placement.
We associate an appearance embedding with each mesh vertex and interpolate these embeddings across the mesh surface to predict Gaussian attributes. 
This design maintains robustness even under occlusions and rapid hand motions.
To better capture dynamic hand motion in egocentric videos, we further incorporate temporal convolution layers to aggregate visual features from adjacent frames for hand pose estimation.
Combined with our representation, these temporally aggregated features enable our feed-forward network to produce stable reconstructions of hand geometry and appearance while capturing dynamic hand motion in challenging egocentric videos.

We evaluate our framework on the H2O~\cite{Kwon_2021_ICCV} and ARCTIC~\cite{fan2023arctic} datasets, which contain challenging egocentric hand--object interactions, occlusions, and dynamic bimanual motions.
We evaluate both 4D hand reconstruction quality and hand pose accuracy.
We compare our method with representative per-video neural rendering baselines, including HandNeRF++, GaussianHand, and EVA, under the same evaluation protocol.
The experimental results demonstrate the effectiveness of our method in reconstructing hand geometry and appearance across challenging sequences.
In particular, ARCTIC is not included in training, and the results show that our pretrained model can efficiently adapt to unseen sequences from a different dataset.
Furthermore, image supervision through differentiable rendering provides an additional signal for slightly refining the initial hand poses by aligning the rendered hands with the input observations, without requiring ground-truth 3D hand poses for the target sequences.

To the best of our knowledge, prior work has not explored feed-forward 3D Gaussian-based egocentric 4D hand reconstruction.
\section{Related Work}
\label{sec:relwork}
In this section, we review prior works closely aligned with our approach: 3D hand reconstruction using templates or 3D Gaussians (\sref{sec:relwork:hand}), feed-forward approaches for 3D human reconstruction (\sref{sec:relwork:human}), and dynamic 3D scene reconstruction (\sref{sec:relwork:dynamic}).

\subsection{3D Hand Reconstruction}
\label{sec:relwork:hand}
Template-based 3D hand pose estimation methods that leverage MANO~\cite{MANO} have established the foundation for monocular 3D hand estimation. Early approaches include \cite{rhoi2020}, which uses contact priors, and \cite{rong2021frankmocap}, which applies encoder-decoder networks. 
For occlusions, \cite{moon2023interwild} introduces normalized scale processing with wrist features. 
Recent advances employ transformers. 
\cite{pavlakos2024reconstructing,potamias2025wilor} scales up with a transformer backbone while \cite{handformer} implements a multi-task progressive architecture. 
However, as these methods process frames independently, their predictions may exhibit inter-frame variations when applied to videos.
To address this, \cite{duran2024hmp, hasson2019learning, liu2021semi, spurr2021adversarial, yang2020seqhand, ziani2022tempclr} leverage video-based approaches for hand pose estimation. 
Other studies target two-hand reconstruction, which is crucial for modeling bimanual manipulation~\cite{yu2023acr, ren2023decoupled, li2022interacting}.
More recent efforts employ SLAM-aided pipelines~\cite{yu2024dynhamr, zhang2025hawor} or multi-frame transformer architectures with cross-view attention to jointly estimate global hand pose and camera trajectory~\cite{ye2025predicting}.
Meanwhile, several works focus on generating hand avatars~\cite{karunratanakul2023harp, moon2024authentic, zheng2024ohta, hu2024expressive, chen2025interactavatar, chen2023hand, huang2024learning,qian2020html,potamias2023handy}. However, these methods are either limited to 3D reconstruction without temporal cues or rely on off-the-shelf hand pose estimators.

Implicit representations have emerged to overcome the limitations of template-based approaches. \cite{mundra2023livehand} combines MANO-guided sampling with a canonicalized texture space, and \cite{guo2023handnerf,guo2025handnerf++} proposes a pose-driven canonical NeRF. Recent works adopt 3DGS in various forms. \cite{pokhariya2023manus} anchors articulated Gaussians to MANO joints, \cite{sun2025jghandjointdrivenanimatablehand} introduces joint-driven warping, and \cite{dong2025handsplatembeddingdrivengaussiansplatting} uses implicit geometry embeddings. \cite{gaussianhand} integrates hierarchical Gaussian blend shapes and neural residual skinning on MANO surfaces. However, despite incorporating hand priors, these methods still rely on multi-view inputs. For monocular videos, \cite{fan2024hold} reconstructs hands using implicit SDFs with pose estimators, while \cite{on2025bigsbimanualcategoryagnosticinteraction} leverages diffusion-prior SDS loss. 

\subsection{Feed-Forward 3D Human Reconstruction}
\label{sec:relwork:human}
Recent feed-forward approaches reduce the need for costly per-scene optimization, enabling efficient reconstruction and adaptation on unseen inputs. 
Several works adapt the feed-forward Gaussian splatting for human reconstruction. 
GPS-Gaussian~\cite{zheng2024gpsgaussian} achieves high-quality reconstruction without explicit templates but requires ground-truth depth and multi-view supervision during training. 
GHG~\cite{kwon2024ghg} employs mesh templates to guide Gaussian placement but relies on ground-truth template poses and texture maps with multi-view inputs.
LIFe-GoM~\cite{wen2025lifegom} learns Gaussian on mesh surfaces, but focuses on capturing fine-grained details with iterative refinement. 
RoGSplat~\cite{RoGSplat2025CVPR} addresses robustness to inaccurate input poses but requires a multi-view input and depth alignment process.

To address the challenge of acquiring multi-view data, which often requires specialized capture setups and controlled environments, recent works attempt monocular approaches. 
HumanSplat~\cite{pan2024humansplat} and IDOL~\cite{zhuang2024idolinstant} utilize diffusion models to synthesize multi-view supervision from a single image, while GUAVA~\cite{GUAVA} obtains UV Gaussian by inverse texture mapping of feature maps obtained from the vision foundation model. However, these methods focus on static 3D reconstruction from images, whereas we focus on dynamic reconstruction from video. 

\subsection{Feed-Forward Dynamic 3D Reconstruction}
\label{sec:relwork:dynamic}
Recent works extend feed-forward methods to video sequences. Forge4D~\cite{hu2025forge4dfeedforward4dhuman} presents the first feed-forward model for dynamic human reconstruction but requires multi-view video inputs. L4GM~\cite{ren2024l4gm} uses monocular video but trains with multi-view supervision synthesized via video diffusion models, introducing computational overhead and potential artifacts. 4DGT~\cite{xu20254dgt} predicts 4D Gaussians from general monocular videos without template priors but lacks specialized modeling for articulated hands. However, prior feed-forward methods have not specifically addressed dynamic hand reconstruction in the egocentric video setting.

\begin{figure*}[tb]
\centering
\includegraphics[width=1\linewidth]{figs/overview_compressed.pdf}
\caption{\textbf{Framework Overview.}
(a) Given an input sequence, a pretrained image encoder extracts per-frame visual tokens, which are processed by temporal convolution layers for hand pose, translation, and vertex embedding prediction.
The predicted MANO mesh and vertex embeddings are used to construct a 3D Gaussian hand for differentiable rendering.
(b) Gaussians are sampled over the mesh surface, and their embeddings are obtained by barycentric interpolation of mesh-aligned vertex embeddings.
Lightweight MLPs predict Gaussian color, scale, and bounded mesh-relative adjustments.
(c) For each visual token, features from adjacent frames are processed by a 1D convolution along the temporal axis.
This layer reduces the feature dimension and incorporates temporal context, providing temporally aggregated features. 
}
\label{fig:main:overview}
\end{figure*}

\section{Method}
\label{sec:method}

In this section, we present our feed-forward framework for egocentric hand reconstruction. 
We begin by defining the task and providing preliminaries on feed-forward 3D Gaussian Splatting (\sref{sec:method:task}). 
We then describe our mesh-guided representation (\sref{sec:method:mesh}), followed by the overall framework (\sref{sec:method:framework}), temporal convolution layer (\sref{sec:temporal}), and loss functions (\sref{sec:loss}).

\subsection{Task Definition}
\label{sec:method:task}

Our task is to reconstruct dynamic hands from single-view egocentric videos. 
Given an input video sequence $\{\mathbf{I}_t\}_{t=1}^T$ captured from an egocentric viewpoint, where $T$ denotes the number of frames, we aim to predict 3D Gaussians that accurately represent the hand geometry and appearance. 
We describe our framework in the following sections.

\paragraph{Preliminary: feed-forward 3DGS.}
Unlike standard 3DGS that optimizes 3D primitives per scene, feed-forward 3DGS predicts 3D Gaussian parameters directly from input images. This enables efficient training across multiple instances and generalization to unseen scenes.

Typically, this approach consists of an encoder $\text{Enc}(\cdot)$ and decoder networks. 
The encoder extracts per-frame features $\mathbf{f}_t = \text{Enc}(\mathbf{I}_t)$ from input frames $\mathbf{I}_t$ independently.

The decoder networks, typically implemented as lightweight MLPs, then map these features to Gaussian parameters $\{\mu, r, s, \sigma, c\}$, where $\mu, r, s \in \mathbb{R}^3$ represent position, rotation, and scale, respectively; $\sigma$ is opacity, and $c \in \mathbb{R}^3$ is color.
Following~\cite{kerbl3Dgaussians}, the predicted 3D Gaussians are projected to 2D space and rendered via alpha-compositing with differentiable rasterization.

\subsection{Mesh-Guided Representation for Hands}
\label{sec:method:mesh}

Unlike general objects, hands exhibit consistent articulated structure across instances, which can be effectively modeled using a hand template. Therefore, we design a mesh-guided representation to leverage this structural prior.

Briefly, 1) we predict MANO mesh, 2) predict mesh-aligned vertex embeddings, 3) sample Gaussians over the mesh surface and obtain their embeddings by barycentric interpolation, 4) define their base positions and orientations from the mesh geometry, and 5) decode the interpolated embeddings to predict color, scale, and bounded mesh-relative adjustments.

\paragraph{Mesh-aligned vertex embeddings.}
First, we predict hand mesh vertices using a hand template (MANO~\cite{MANO}) from the encoder features $\mathbf{f}$.
We also predict a set of mesh-aligned vertex embeddings through a lightweight MLP:
\begin{equation}
    \mathbf{E} = \text{MLP}_{\text{emb}}(\mathbf{f}),
\end{equation}
where $\mathbf{E}=\{e_i\}_{i=1}^{N_v}$, $N_v$ is the number of mesh vertices, and $e_i$ denotes the embedding associated with vertex $v_i$.
These vertex embeddings serve as conditional latents for predicting Gaussian parameters.

Since the embeddings are defined only at the mesh vertices, they cannot directly provide an embedding for each Gaussian sampled within a triangular face. 
We therefore distribute a fixed number of Gaussians over the mesh surface by sampling triangular faces in proportion to their surface areas.
For each sampled Gaussian, we obtain its embedding $\tilde{e}$ through barycentric interpolation of the three neighboring vertex embeddings:
\begin{equation}
    \tilde{e} = \text{Interp}(e_i, e_j, e_k),
\end{equation}
where $(e_i, e_j, e_k)$ are the embeddings of the three triangle vertices. 
This provides each sampled Gaussian with a mesh-aligned embedding derived from its neighboring vertex embeddings.

The interpolated embedding $\tilde{\mathbf{e}}$ is decoded by three lightweight MLP decoders, $\text{MLP}_{\mathrm{color}}$, $\text{MLP}_{\mathrm{scale}}$, and $\text{MLP}_{\mathrm{adjust}}$, which predict color, Gaussian scale, and mesh-relative Gaussian adjustments, respectively.
The color decoder predicts zeroth-order spherical harmonic coefficients $c \in \mathbb{R}^3$, and the scale decoder predicts anisotropic scale $s \in \mathbb{R}^3$. 
The Gaussian adjustment decoder adjusts the barycentric position, normal displacement, and Gaussian orientation within the local tangent plane.
These adjustments are bounded to keep the Gaussians closely associated with the underlying mesh surface while allowing them to better capture local appearance and geometric details.
All decoders use a single 64-dimensional hidden layer. 
Further details are provided in the supplementary material.

\paragraph{Mesh-relative Gaussian adjustment.}
For each Gaussian, the initial barycentric coordinates $\mathbf{b}$ are kept fixed after surface sampling and define its base position on the associated mesh face.
The adjustment decoder predicts residuals $\Delta \mathbf{b}$, $\Delta d$, and $\Delta \phi$ for the barycentric position, normal displacement, and in-plane rotation, respectively.
We update the barycentric coordinates as $\mathbf{b}'=\mathrm{Softmax}(\log \mathbf{b}+\Delta \mathbf{b})$, which preserves the initial position when $\Delta\mathbf{b}=0$ and keeps the Gaussian inside the face.
The position is then displaced by $\Delta d$ along the face normal, while $\Delta\phi$ refines the orientation within the local tangent plane.
This preserves the association between each Gaussian and its underlying mesh while allowing its position and orientation to be refined.

\subsection{Overall Framework}
\label{sec:method:framework}

\cref{fig:main:overview} illustrates our overall pipeline. Given an input video frame $I_t$, we first extract visual tokens using a pretrained vision backbone. These features are processed through a temporal convolution layer to compress and incorporate temporal context across frames, which we detail in Section~\ref{sec:temporal}.

A pose transformer~\cite{vaswani2017attention} then predicts the MANO pose parameters from temporally aggregated features:
\begin{equation}
    \{\bm{\beta}, \bm{\theta}\} = \text{Transformer}_{\text{MANO}}(\bar{\mathbf{f}}_t).
\end{equation}
where $\bm{\beta} \in \mathbb{R}^{10}$ and $\bm{\theta} =[\bm{\theta}_{\text{root}};\bm{\theta}_{\text{hand}}] \in \mathbb{R}^{48}$ denote shape and pose parameters, respectively.
In parallel, the translation $\bm{\tau}_t \in \mathbb{R}^{3}$ is predicted by a shallow MLP
with two 512-dimensional hidden layers and ReLU activations.
Note that $\bar{\mathbf{f}}_t$ represents temporally aggregated features from adjacent frames, as detailed in \sref{sec:temporal}. These parameters are fed into the MANO model to extract hand mesh vertices $\mathbf{v}$. To establish reliable initial geometry, we supervise the predicted vertices using pseudo ground truth obtained from an off-the-shelf pose estimator~\cite{potamias2025wilor}.

Simultaneously, we predict per-vertex embeddings and interpolate them to obtain additional Gaussian embeddings across the mesh surface. 
The interpolated embeddings are decoded to predict Gaussian color, scale, and mesh-relative adjustments. 
Base positions and orientations are derived from the mesh geometry and further refined by the adjustment decoder.

The predicted 3D Gaussians are rendered into an image, enabling image supervision. 
This image-level supervision jointly refines both the predicted mesh vertices and Gaussian parameters during training.

\subsection{Temporal Convolution Layer}
\label{sec:temporal}

As described in the preliminary section, standard feed-forward approaches extract per-frame features independently. In our implementation, this produces features $\mathbf{f}_{t} \in \mathbb{R}^{N \times D}$, where $N$ is the number of visual tokens and $D$ is the feature dimension. 

Since the visual backbone processes each frame independently, the resulting representations may exhibit inter-frame variations. We therefore apply a temporal convolution layer to jointly process features from consecutive frames.
For each frame $t$, we aggregate features within a temporal window of size $2k+1$:
\begin{equation}
    \bar{\mathbf{f}}_{t} = \text{TC}(\mathbf{f}_{t-k}, \ldots, \mathbf{f}_t, \ldots, \mathbf{f}_{t+k}),
    \label{eq:temporal_conv}
\end{equation}
where $\text{TC}$ denotes the temporal convolution operation. This produces temporally consistent features $\bar{\mathbf{f}}_t \in \mathbb{R}^{N \times D'}$ with reduced dimension $D' < D$, which are used for subsequent MANO and embedding predictions.
These temporally aggregated features help maintain geometric consistency across frames while capturing hand motion. 

\subsection{Training Objectives}
\label{sec:loss}

\paragraph{Pose supervision.}
We supervise the predicted MANO parameters and mesh vertices using pseudo ground truth obtained from an off-the-shelf hand pose estimator:
\begin{equation}
\begin{aligned}
\mathcal{L}_{\mathrm{pose}}
={}&
\lambda_{\mathrm{vert}}
\left\|
\mathbf{v}-\mathbf{v}_{\mathrm{pseudo}}
\right\|_1
+
\lambda_{\theta}
d_{\mathrm{chord}}
\left(
\bm{\theta},\bm{\theta}_{\mathrm{pseudo}}
\right) \\
&+
\lambda_{\beta}
\left\|
\bm{\beta}-\bm{\beta}_{\mathrm{pseudo}}
\right\|_2^2
+
\lambda_{\tau}
\left\|
\bm{\tau}-\bm{\tau}_{\mathrm{pseudo}}
\right\|_1 ,
\end{aligned}
\end{equation}
where $d_{\mathrm{chord}}$ denotes the squared chordal distance between the corresponding rotation matrices.

\paragraph{Image supervision.}
We supervise the rendered hand using an $\ell_1$ reconstruction loss
and structural dissimilarity:
\begin{equation}
\mathcal{L}_{\mathrm{img}}
=
\left\|
\mathbf{M}_t\odot
\left(\hat{\mathbf{I}_t}-\mathbf{I}_t\right)
\right\|_1
+
\lambda_{\mathrm{ssim}}
\mathcal{L}_{\mathrm{D\text{-}SSIM}}
\left(\hat{\mathbf{I}_t},\mathbf{I}_t;\mathbf{M}_t\right),
\end{equation}
where $\mathbf{M}_t$ denotes the valid hand-region mask at a frame $t$.

\paragraph{Regularization.}
We employ a local Wasserstein distance~\cite{balle2025good} loss to improve appearance details. We also apply an alpha leakage loss, which penalizes the proportion of rendered alpha that falls outside the hand mask. This suppresses rendering artifacts in background regions without directly constraining individual Gaussian opacities. The regularization loss is defined as
\begin{equation}
\mathcal{L}_{\mathrm{reg}}
=
\lambda_{\mathrm{WD}}\mathcal{L}_{\mathrm{WD}}
+
\lambda_{\alpha}\mathcal{L}_{\alpha}.
\end{equation}

\paragraph{Total loss.}
Our final training objective is
\begin{equation}
\mathcal{L}
=
\mathcal{L}_{\mathrm{pose}}
+
\mathcal{L}_{\mathrm{img}}
+
\mathcal{L}_{\mathrm{reg}}.
\end{equation}

\section{Experiments}
\label{sec:exp}

In this section, we first explain experimental details including datasets, baselines, and metrics (\sref{exp:details}). Then, we demonstrate the effectiveness of our approach on reconstruction performance (\sref{exp:results}). Finally, we conduct analyses of our method and ablation studies (\sref{exp:anal}).

\subsection{Experiment Details}
\label{exp:details}

\paragraph{Datasets.}
For pose pretraining and joint training, we use H2O~\cite{Kwon_2021_ICCV}, TACO~\cite{liu2024taco}, and HoloAssist~\cite{wang2023holoassist}. 
For appearance pretraining, we additionally incorporate PALM~\cite{fan2026palm} to increase appearance diversity.

We evaluate our method on H2O and ARCTIC~\cite{fan2023arctic}.
H2O captures egocentric hand-object interactions with RGB-D sequences, camera poses, and 3D hand poses. We use sequences 0--7 from \texttt{subject4/h1} as the test set and all other sequences for training, using only RGB frames from the egocentric viewpoint. 
For the baselines, every tenth frame is held out for evaluation, while the remaining frames are used to fit a separate model for each test sequence.
Our model is pretrained without any frames from the test sequences and uses only the non-held-out frames for short test-time optimization.
ARCTIC captures two-handed manipulation of articulated objects using multi-view cameras and a MoCap system. 
ARCTIC features particularly challenging scenarios with fast hand motions and complex interactions. 
ARCTIC is unseen by our offline training data and is used to evaluate cross-dataset generalization with test-time adaptation.
We evaluate four \texttt{grab} sequences and four \texttt{use} sequences from \texttt{s01}. 
The baselines follow the same per-sequence fitting and holdout protocol as on H2O.
Further details regarding the datasets are provided in the supplementary material.

\begin{figure*}[tb!]
\centering
\includegraphics[width=1\linewidth]{figs/h2o-3dv2.pdf}
\caption{\textbf{Reconstruction Fidelity on the H2O dataset.}
The baselines exhibit stripe-like artifacts, global color mismatches, or Gaussian artifacts around the hand boundaries.
In contrast, Hand-4DGS maintains stable mesh-guided geometry and reconstructs hand appearance with fewer visual artifacts and cleaner hand boundaries.
}
\label{fig:main:h2o}
\end{figure*}

\paragraph{Baselines.}
We compare our method with HandNeRF++~\cite{guo2023handnerf,guo2025handnerf++}, GaussianHand~\cite{huang2024learning}, and EVA~\cite{hu2024expressive}, which represent three relevant categories of monocular hand reconstruction methods.
We select HandNeRF++ as a radiance-field-based hand reconstruction baseline and GaussianHand as a generalizable hand-avatar baseline. 
For GaussianHand, we follow its one-shot fitting protocol using 100 optimization steps. 
We select EVA as an optimization-based monocular Gaussian baseline and adapt it to hands by replacing its body template with MANO. 
We use the official implementations and configurations for all baselines. For training, all methods use hand poses and hand masks estimated by WiLoR~\cite{potamias2025wilor}, along with SAM3~\cite{carion2026sam} masks to handle occluded regions.

\paragraph{Metrics.}
We evaluate reconstruction fidelity and hand pose accuracy.
For reconstruction quality, we report Peak Signal-to-Noise Ratio (PSNR) and Structural Similarity Index Measure (SSIM).
For each frame, we composite the ground-truth hand over a white background using the ground-truth visible-hand mask and compare it with the full white-background rendering.
To penalize artifacts outside the ground-truth hand boundaries, we define the evaluation region by dilating the visible-hand mask by 16 pixels.
We use SAM3 to obtain object masks and exclude regions where the hands are occluded by objects from the evaluation.
Per-frame results are averaged over the evaluation set, and the same regions are used for all methods.

For hand pose accuracy, we report Root-Aligned Mean Per-Joint Position Error
(RA-MPJPE) and Absolute Mean Per-Joint Position Error (Abs-MPJPE). For
RA-MPJPE, we align the predicted and ground-truth 3D joints at the root joint
for each hand and frame, and then compute their mean Euclidean distance. This
metric evaluates articulated hand pose independently of global hand
translation.

For Abs-MPJPE, we transform the predicted joints from camera coordinates to
world coordinates at each frame using the ground-truth camera extrinsics and
directly compare them with the ground-truth joints without root alignment.
Abs-MPJPE therefore evaluates both articulated pose and absolute hand
translation. Both metrics are reported in millimeters, where lower values
indicate better accuracy.

\paragraph{Training and implementation details.}
We train our model in two stages.
In the first stage, we independently pretrain the pose and translation branches for 300K steps using $\mathcal{L}_{\mathrm{pose}}$, and the appearance branch for 20K steps using $\mathcal{L}_{\mathrm{img}}+\mathcal{L}_{\mathrm{reg}}$.
In the second stage, we jointly optimize the pose, translation, and appearance components for 20K steps using the full objective in Eq.~(8). We use a batch size of 4 during training.

We initialize 15K Gaussians per hand by uniformly sampling the MANO surface, where each face is selected in proportion to its surface area.
The sampled Gaussians remain associated with the initial mesh faces, and Gaussian splitting and pruning are disabled.
We use 16-dimensional per-vertex embeddings and obtain each Gaussian embedding through barycentric interpolation.

Image supervision is applied at the native image resolution within valid hand regions.
We use the intersection of the WiLoR hand region and the SAM3 hand mask to reduce supervision from occluding objects and background regions.
We set the D-SSIM weight to $0.1$, the local WD weight to $0.005$, and the alpha leakage weight to $0.1$.

At sequence boundaries, temporal inputs are constructed using reflection padding without mixing frames from different sequences.
During evaluation, we perform 200 steps of per-sequence test-time optimization (TTO) with a batch size of 1 using the non-held-out frames. 
Additional optimization schedules, loss weights, and architectural details are provided in the supplementary material.

\subsection{Experiment Results}
\label{exp:results}

\begin{table}[t]
  \caption{\textbf{Quantitative comparison on the H2O dataset.}
HandNeRF++ and GaussianHand keep the input poses fixed.
EVA allows pose refinement during optimization, but its estimated poses show negligible improvement over the initialization.
In contrast, Hand-4DGS jointly refines pose and appearance, achieving a slight improvement in pose accuracy over the input poses.}
  \label{tab:h2o_results}
  \centering
  \setlength{\tabcolsep}{4pt}
  \resizebox{1\linewidth}{!}{
    \begin{tabular}{@{}lcccc@{}}
      \toprule
      Method
      & PSNR $\uparrow$
      & SSIM $\uparrow$
      & RA-MPJPE $\downarrow$
      & Abs-MPJPE $\downarrow$ \\
      \midrule
      HandNeRF++
      & 13.61
      & 0.671
      & 16.896
      & 29.988 \\

      GaussianHand
      & 12.22
      & 0.604
      & 16.896
      & 29.988 \\

      EVA
      & 12.40
      & 0.643
      & 16.899
      & 29.986 \\

      Hand-4DGS
      & \textbf{15.48}
      & \textbf{0.693}
      & \textbf{15.708}
      & \textbf{28.025} \\
      \bottomrule
    \end{tabular}
  }
\end{table}

\paragraph{Reconstruction fidelity.}
\label{exp:results_recon}

We compare the reconstruction quality of our method with existing baselines on H2O.
The quantitative results are reported in \cref{tab:h2o_results}, and qualitative comparisons are shown in \cref{fig:main:h2o}.

All baselines use MANO to provide an underlying hand structure, which helps them preserve the overall hand shape.
However, each method exhibits distinct artifacts in its reconstructed hands.
HandNeRF++ uses a mesh-guided neural radiance field to model articulated hand geometry and appearance. 
While it preserves the overall hand structure, its renderings exhibit stripe-like artifacts.
GaussianHand maintains the overall hand geometry through MANO guidance, but its test-time optimization struggles to recover the target hand appearance.
EVA uses a deformable canonical Gaussian avatar, but its results show blurred hand details and Gaussian artifacts outside the hand boundaries.
Unlike approaches that use MANO primarily as an underlying geometric or deformation prior, our mesh-guided representation explicitly associates Gaussians with the mesh surface. 
Their appearance and local adjustments are derived from mesh-aligned vertex embeddings through barycentric interpolation. 
This design maintains stable hand geometry while modeling appearance variations across the sequence.

\paragraph{Cross-dataset generalization.}   
To evaluate cross-dataset generalization, we report results on ARCTIC in \cref{tab:arctic_results}, which is not used during either pretraining or joint training.
Compared with H2O, the selected ARCTIC sequences contain faster hand motions and more severe occlusions, making appearance reconstruction particularly challenging for the baselines. 
Despite this domain shift, our method adapts to each sequence in approximately one minute and achieves competitive reconstruction performance. 
Qualitative results are provided in the supplementary material.

To further evaluate our method on in-the-wild videos with more complex and diverse backgrounds, \fref{fig:in-the-wild} presents qualitative examples from the EPIC-KITCHENS~\cite{damen2022rescaling} and ADL~\cite{pirsiavash2012detecting} datasets.
Although neither dataset is included in training, the results indicate that our model can adapt to unseen scenes through efficient TTO.

\begin{figure*}[h]
\centering
\includegraphics[width=0.9\linewidth]{figs/in-the-wild.pdf}
\caption{\textbf{In-the-Wild Video Results.} 
For each example, the input frame and corresponding hand reconstruction are shown on the left and right, respectively. 
The top row shows results from ADL, while the bottom two rows show results from EPIC-KITCHENS.
}
\label{fig:in-the-wild}
\end{figure*}

\subsection{Analysis}
\label{exp:anal}

\begin{table}[t]
\caption{\textbf{Quantitative comparison on ARCTIC.}
The baselines are optimized separately for each target sequence, whereas Hand-4DGS uses a model pretrained without ARCTIC and performs only short test-time optimization.
Despite the challenging motions and occlusions, Hand-4DGS produces plausible reconstructions with substantially less optimization time.}
\label{tab:arctic_results}
\centering
\setlength{\tabcolsep}{2pt}
\resizebox{\linewidth}{!}{
\begin{tabular}{@{}lccccc@{}}
\toprule
Method
& PSNR $\uparrow$
& SSIM $\uparrow$
& RA-MPJPE $\downarrow$
& Abs-MPJPE $\downarrow$
& Optim. Time $\downarrow$ \\
\midrule
HandNeRF++
& 12.20
& 0.568
& 14.340
& 32.153
& 1 h 37 min \\

GaussianHand
& 11.91
& 0.553
& 14.340
& 32.153
& 1 min 59 sec \\

EVA
& 11.05
& 0.5458
& \textbf{14.326}
& 32.130
& 11 min 3 sec \\

Hand-4DGS
& \textbf{12.98}
& \textbf{0.574}
& 15.085
& \textbf{29.197}
& \textbf{51 sec} \\
\bottomrule
\end{tabular}
}
\end{table}

\paragraph{Analysis of temporal window size.}
\label{exp:results_tc}
The pose and appearance branches use separate temporal convolution layers.
We therefore evaluate the window size separately for the two branches in \cref{tab:window_size}.
For pose prediction, a three-frame window ($k=1$) achieves the highest accuracy, indicating that temporal context from adjacent frames benefits pose estimation. 
In contrast, appearance prediction performs best without temporal aggregation ($k=0$), suggesting that temporal convolution can discard frame-specific appearance information. 
To isolate appearance reconstruction from pose estimation errors, we provide ground-truth hand poses as input for the appearance evaluation.
Based on these results, we use $k=1$ for the pose branch and $k=0$ for the appearance branch.

\begin{table}[t]
  \caption{\textbf{Experiments on the window size.}
  We evaluate the temporal window separately for pose and appearance
prediction. We use ground-truth hand poses for appearance evaluation to isolate appearance reconstruction from pose errors.}
  \label{tab:window_size}
  \centering
  \setlength{\tabcolsep}{2pt}
  \resizebox{1\linewidth}{!}{
    \begin{tabular}{@{}l|cc|cc@{}}
      \toprule
      & \multicolumn{2}{c|}{Pose}
      & \multicolumn{2}{c}{Appearance} \\
      Window Size ($k$)
      & RA-MPJPE $\downarrow$
      & Abs-MPJPE $\downarrow$
      & PSNR $\uparrow$
      & SSIM $\uparrow$ \\
      \midrule
      1 ($k=0$)
      & 17.964
      & 31.055
      & \textbf{16.65}
      & \textbf{0.728} \\

      3 ($k=1$)
      & \textbf{16.996}
      & \textbf{28.793}
      & 16.53
      & 0.722 \\

      5 ($k=2$)
      & 17.693
      & 29.442
      & 16.45
      & 0.721 \\
      \bottomrule
    \end{tabular}
  }
\end{table}

\paragraph{Ablation of appearance modeling.}
We evaluate the contribution of the Gaussian adjustment decoder and local WD regularization to reconstruction quality.
As shown in \cref{tab:component_ablation}, removing either component degrades the quantitative performance.
Without the adjustment decoder, the Gaussians do not fully cover regions near the hand boundaries, and visible gaps remain.
Without local WD regularization, fine textures and subtle shading are weakened, and the reconstructed appearance becomes smoother.
These observations indicate that the adjustment decoder improves boundary coverage,  while local WD helps model appearance details and local shading.
Corresponding qualitative examples are provided in the supplementary material.

\begin{table}[t]
  \caption{\textbf{Ablation study of the proposed components.}
  We evaluate the Gaussian adjustment decoder and local WD regularization.}
  \label{tab:component_ablation}
  \centering
  \setlength{\tabcolsep}{6pt}
  \resizebox{0.9\linewidth}{!}{
  \begin{tabular}{@{}lcc@{}}
    \toprule
    Configuration
    & PSNR $\uparrow$
    & SSIM $\uparrow$ \\
    \midrule
    Full
    & \textbf{17.15}
    & \textbf{0.731} \\
    
    w/o Gaussian Adjustment decoder
    & 17.05
    & 0.729 \\
    
    w/o Local WD
    & 16.92
    & 0.729 \\
    \bottomrule
  \end{tabular}
  }
\end{table}

\section{Conclusion}
\label{sec:conclusion}
In this paper, we presented Hand-4DGS, a framework for reconstructing dynamic two-hand geometry and appearance from monocular egocentric videos. 
Our method combines a mesh-guided Gaussian representation with temporally aggregated visual features. 
Per-vertex embeddings and their barycentric interpolation provide structured appearance features across the hand surface, while lightweight Gaussian decoders model color, scale, and local adjustments. 
The model jointly predicts hand pose and appearance and can efficiently adapt to previously unseen video sequences.

Our experiments demonstrate that Hand-4DGS produces stable hand reconstructions across videos containing complex interactions, occlusions, and dynamic hand motions. 
The model also slightly improves the initial hand poses through joint optimization with image supervision. 
Furthermore, the results on in-the-wild videos show its ability to adapt to scenes outside the training datasets.

\paragraph{Limitations.}
Our method does not always recover subtle appearance details, such as visible veins and fine skin wrinkles.
Nevertheless, our main strength lies in reconstructing hands across diverse videos with complex and dynamic motions, for which stable geometry and appearance are particularly challenging to maintain.
Future work may improve the representation of such high-frequency appearance details and extend the framework to jointly model hands and interacting objects.

{
    \small
    \bibliographystyle{ieeenat_fullname}
    \bibliography{hand4dgs_refs}
}

\clearpage
\setcounter{page}{1}
\maketitlesupplementary

\renewcommand{\thesection}{\Alph{section}}
\setcounter{section}{0}
\renewcommand{\thetable}{S\arabic{table}}
\renewcommand{\thefigure}{S\arabic{figure}}
\setcounter{figure}{0}
\setcounter{table}{0}

This supplementary material provides additional architectural analyses,
implementation details, dataset and evaluation protocols, and novel-view
rendering results.

\section{Qualitative Results on ARCTIC}
\label{sec:supp_arctic}

Figure~\ref{fig:supp_arctic} presents qualitative comparisons on ARCTIC, which is not used during the offline training of Hand-4DGS.
HandNeRF++ preserves the overall hand structure but exhibits noticeable appearance artifacts. 
GaussianHand struggles to recover the target appearance, resulting in inaccurate colors and textures. 
EVA produces blurred hand details and floaters around the hand boundaries. 
Despite the cross-dataset domain shift, Hand-4DGS produces plausible hand geometry and appearance after short test-time optimization.

\begin{figure*}[t]
    \centering
    \includegraphics[width=\textwidth]{figs/arctic.pdf}
    \caption{\textbf{Qualitative comparisons on ARCTIC.}
    HandNeRF++ exhibits appearance artifacts, GaussianHand struggles to recover the target appearance, and EVA produces blurred hand details and floaters around the hand boundaries. 
    Despite the cross-dataset domain shift, Hand-4DGS produces plausible hand geometry and appearance. 
    ARCTIC is not used during the offline training of Hand-4DGS.}
    \label{fig:supp_arctic}
\end{figure*}

\section{Qualitative Analysis of Appearance Components}
\label{sec:supp_appearance_analysis}

\begin{figure}[t]
    \centering
    \includegraphics[width=\linewidth]{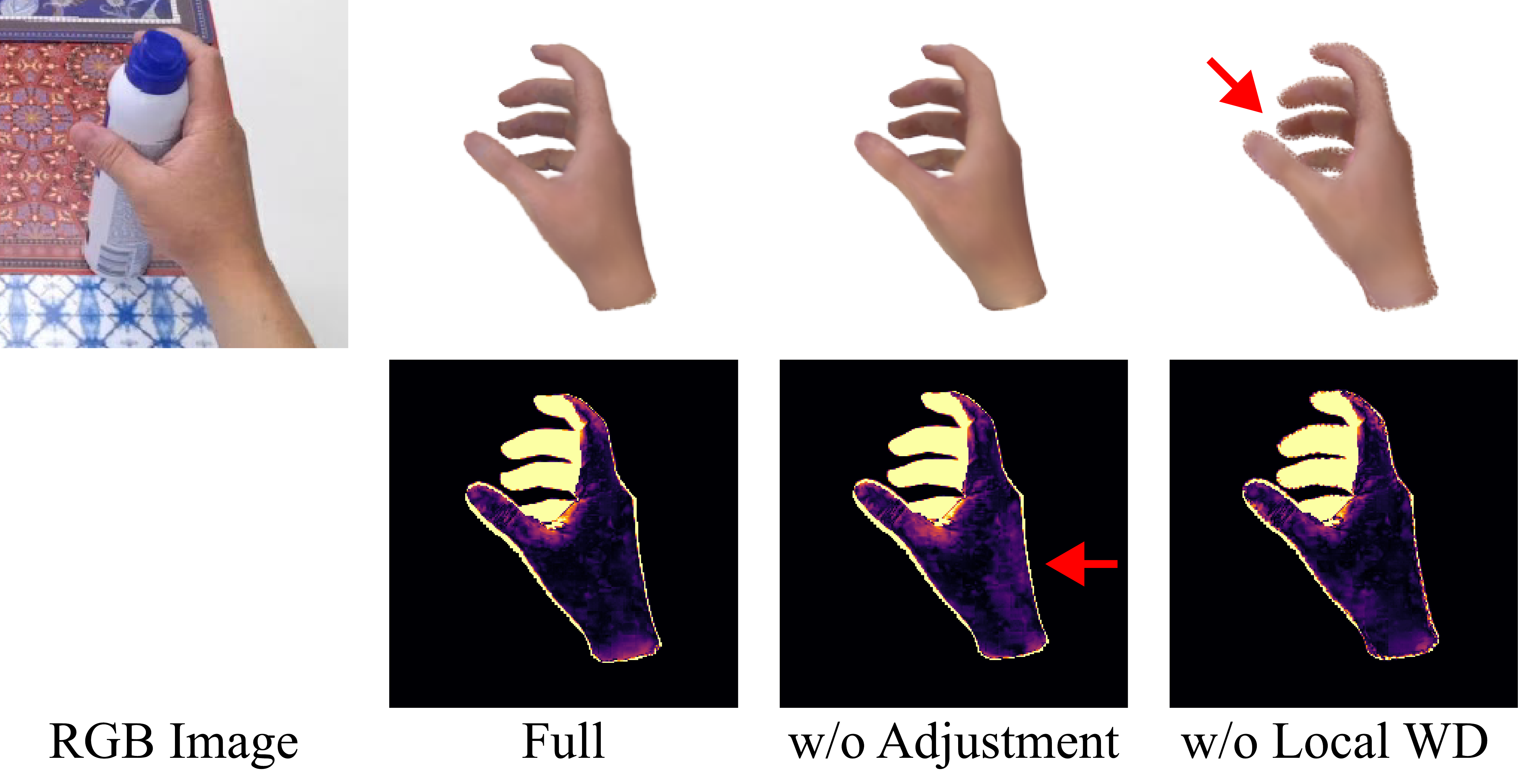}
    \caption{\textbf{Qualitative analysis of appearance components.}
The top row shows the rendered hands, while the bottom row shows the corresponding pixel-wise error maps with respect to the target image.
Removing the mesh-relative Gaussian adjustment produces irregular artifacts around the finger boundaries, whereas removing the local WD loss increases errors across broad surface regions with gradual shading variations.}
    \label{fig:supp_appearance_analysis}
\end{figure}

Figure~\ref{fig:supp_appearance_analysis} presents qualitative comparisons of the components used for appearance reconstruction. 
Without the mesh-relative Gaussian adjustment, the surface-sampled Gaussians cannot adapt their positions and orientations to the local hand geometry, producing irregular artifacts near the hand boundaries. 
Without the local WD loss, reconstruction errors increase across broad surface regions, particularly in areas with gradual shading variations. These results suggest that the Gaussian adjustment improves rendering near boundaries, while the local WD loss helps preserve local appearance and shading.

\section{Effect of Joint Training}
\label{sec:supp_joint_training}

We analyze the effect of jointly optimizing the pose, translation, and appearance branches. 
As shown in \cref{tab:supp_joint}, joint training improves PSNR and SSIM and reduces RA-MPJPE compared with separate pretraining. 
These results indicate that joint image and pose supervision improves both appearance reconstruction and root-relative articulated pose.

In contrast, Abs-MPJPE increases after joint training. 
Unlike RA-MPJPE, Abs-MPJPE retains the global root translation and is therefore more sensitive to monocular depth ambiguity. 
The lower RA-MPJPE and higher Abs-MPJPE suggest that this difference mainly arises from global hand translation rather than articulated pose.
Overall, joint training improves appearance fidelity and root-relative pose accuracy, with a modest decrease in absolute hand-placement accuracy. 
Note that no ground-truth 3D hand poses are used during either pretraining or joint training.

\begin{table}[t]
  \caption{\textbf{Effect of joint training.}
  We compare the separately pretrained model before joint optimization with the model after 20K joint-training iterations.}
  \label{tab:supp_joint}
  \centering
  \setlength{\tabcolsep}{3pt}
  \resizebox{\linewidth}{!}{
  \begin{tabular}{@{}lcccc@{}}
    \toprule
    Training stage
    & PSNR $\uparrow$
    & SSIM $\uparrow$
    & RA-MPJPE $\downarrow$
    & Abs-MPJPE $\downarrow$ \\
    \midrule
    Before joint training
    & 13.9836
    & 0.6588
    & 17.2531
    & \textbf{26.0723} \\
    Joint training (20K)
    & \textbf{15.4800}
    & \textbf{0.6930}
    & \textbf{15.7080}
    & 28.0250 \\
    \bottomrule
  \end{tabular}
  }
\end{table}

\section{Novel-View Rendering}
\label{sec:supp_novel}

We evaluate novel-view rendering to examine whether the reconstructed hands maintain coherent appearance beyond the egocentric input viewpoint. 
To separate appearance fidelity from global hand-placement errors, we apply translation-only root alignment between the reconstructed and ground-truth hands before rendering. 
PSNR and SSIM are then computed using the same hand-focused evaluation protocol as in the main paper.

Table~\ref{tab:supp_novel} reports the results averaged over four held-out cameras for sequences 0--7 from \texttt{subject4/h1} in H2O, comprising 2,112 view--frame samples. 
Hand-4DGS achieves the highest PSNR, while GaussianHand achieves the highest SSIM. 
However, as shown in \cref{fig:supp_novelview}, GaussianHand fails to recover the target appearance colors despite preserving the overall hand structure. 
In comparison, Hand-4DGS produces more visually coherent appearance across the held-out viewpoints.

\begin{table}[t]
  \caption{\textbf{Novel-view reconstruction on H2O.}
  We apply translation-only root alignment before rendering to separate
appearance fidelity from global hand-placement errors. 
Results are averaged over four held-out cameras for sequences 0--7 of
\texttt{subject4/h1}.}
  \label{tab:supp_novel}
  \centering
  \setlength{\tabcolsep}{4pt}
  \resizebox{0.5\linewidth}{!}{
  \begin{tabular}{@{}lcc@{}}
    \toprule
    Method & PSNR $\uparrow$ & SSIM $\uparrow$ \\
    \midrule
    HandNeRF++          & 11.9106 & 0.4724 \\
    GaussianHand        & 11.1545 & \textbf{0.5135} \\
    EVA                 & 11.8672 & 0.4391 \\
    Hand-4DGS & \textbf{12.2780} & 0.5047 \\
    \bottomrule
  \end{tabular}
  }
\end{table}

\begin{figure*}[t]
    \centering
    \includegraphics[width=\textwidth]{figs/novel.pdf}
    \caption{\textbf{Qualitative comparison of novel-view rendering on H2O.}
Each group of four rows shows the same timestamp rendered from four held-out cameras.
Before rendering, each reconstructed hand is translation-aligned to the ground-truth root to separate appearance fidelity from global hand-placement errors. 
GaussianHand preserves the overall hand structure but does not accurately recover the target appearance colors. 
Hand-4DGS maintains stable hand geometry and produces more visually coherent appearance across the viewpoints.}
    \label{fig:supp_novelview}
\end{figure*}

\section{Architecture Design and Analysis}
\label{sec:supp_arch}

We analyze the architectural choices for the pose and translation branches.
We first compare Transformer- and MLP-based prediction architectures for
each branch and then evaluate the depth of the selected architectures.
Detailed configurations of the final branches are provided in
\cref{sec:supp_final_arch}.

\subsection{Branch-Specific Prediction Architectures}
\label{sec:supp_branch_design}

We explore Transformer- and MLP-based prediction architectures for both the
pose and translation branches. Both architectures receive temporally
processed visual features but aggregate their spatial information
differently. The MLP flattens the spatial features into a single vector and
directly predicts the output parameters. In contrast, the Transformer
concatenates the spatial visual tokens with learnable task queries and
jointly processes the resulting sequence through self-attention.

For MANO prediction, the Transformer uses seven task queries corresponding
to the root orientation, five finger-pose groups, and MANO shape. Each
finger query represents the three joints of the corresponding finger. The
Transformer variant of the translation branch instead uses a single task
query to jointly predict the crop-camera parameters.

The final output representation at each task-query position is passed to a
lightweight MLP decoder. The pose-query decoders predict the root
orientation, finger-joint rotations, and MANO shape. Following WiLoR, the
translation-query decoder predicts the weak-perspective crop-camera
parameters $(s_\text{cam},t_x,t_y)$, consisting of a scale and two in-plane offsets.

We evaluate the pose architectures using RA-MPJPE and the translation
architectures using root error, defined as the Euclidean distance between
the predicted and ground-truth root positions in absolute coordinates.
Both metrics are reported in millimeters.

\begin{table}[t]
  \caption{\textbf{Comparison of prediction architectures.}
  We compare Transformer- and MLP-based architectures for the pose and
  translation branches while keeping the remaining components unchanged.}
  \label{tab:supp_branch_design}
  \centering
  \setlength{\tabcolsep}{7pt}
  \resizebox{1\linewidth}{!}{
  \begin{tabular}{@{}lc|lc@{}}
    \toprule
    \multicolumn{2}{c|}{Pose branch}
    & \multicolumn{2}{c}{Translation branch} \\
    Architecture
    & RA-MPJPE $\downarrow$
    & Architecture
    & Root Error $\downarrow$ \\
    \midrule
    MLP
    & 46.630
    & Transformer
    & 46.264 \\
    \textbf{Transformer}
    & \textbf{17.039}
    & \textbf{MLP}
    & \textbf{23.871} \\
    \bottomrule
  \end{tabular}
  }
\end{table}

As shown in \cref{tab:supp_branch_design}, the Transformer architecture
achieves substantially lower RA-MPJPE than the MLP architecture for pose
prediction. This result suggests that task-specific queries help the model
aggregate spatial evidence for the root orientation, individual finger
groups, and hand shape.

In contrast, the MLP architecture achieves a considerably lower root error
than the Transformer architecture for translation prediction. We
hypothesize that flattening the spatial features allows the MLP to aggregate
global location and scale information more directly, particularly when
combined with the bounding-box geometry. Based on these results, we use a
Transformer-based pose branch and an MLP-based translation branch in our
final model.

\subsection{Architecture Depth}
\label{sec:supp_arch_depth}

We next evaluate the depth of the selected architectures. For the pose
Transformer, we vary the number of Transformer blocks while keeping the
token dimension, number of attention heads, and feed-forward dimension
fixed. For the translation MLP, we vary the number of hidden layers while
keeping their dimensions fixed.

\begin{table}[t]
  \caption{\textbf{Analysis of architecture depth.}
  We independently vary the number of blocks in the pose Transformer and the number of hidden layers in the translation MLP while keeping the remaining settings unchanged.}
  \label{tab:supp_arch_depth}
  \centering
  \setlength{\tabcolsep}{4pt}
  \resizebox{\linewidth}{!}{
  \begin{tabular}{@{}lc|lc|c@{}}
    \toprule
    \multicolumn{2}{c|}{Pose Transformer}
    & \multicolumn{2}{c|}{Translation MLP}
    & \\
    Blocks
    & RA-MPJPE $\downarrow$
    & Layers
    & Root Error $\downarrow$
    & Abs-MPJPE $\downarrow$ \\
    \midrule
    1
    & 17.483
    & 1
    & 27.986
    & 34.718 \\
    2
    & \textbf{17.039}
    & 2
    & 23.871
    & \textbf{30.882} \\
    3
    & 18.261
    & 3
    & \textbf{23.522}
    & 31.152 \\
    \bottomrule
  \end{tabular}
  }
\end{table}

As shown in \cref{tab:supp_arch_depth}, two Transformer blocks achieve the
lowest RA-MPJPE, while increasing the depth to three blocks does not provide
a further improvement.

For the translation MLP, three hidden layers yield a slightly lower root
error than two hidden layers, whereas two hidden layers achieve the lowest
Abs-MPJPE. Since Abs-MPJPE evaluates all joints in absolute coordinates and
therefore better reflects the absolute placement of the reconstructed hand,
we use two hidden layers in the final translation branch. Overall,
increasing the depth beyond the selected configurations does not provide a
consistent improvement.

\section{Implementation Details}
\label{sec:supp_impl}

\subsection{Input Preprocessing and Feature Extraction}
\label{sec:supp_preprocessing}

We use WiLoR to provide hand-pose inputs for the
baselines and pseudo-labels for supervising the pose branch of our model.
For our model, we additionally use its frozen reconstruction backbone as
the visual encoder.

WiLoR estimates hand pose from localized hand regions rather than directly
processing the full input frame. We follow its original preprocessing
pipeline without modification. Given an input frame, WiLoR first localizes
each hand, constructs a square crop around the detected bounding box, and
resizes it to $256\times256$. Its frozen reconstruction backbone processes
the central $256\times192$ region, obtained by removing 32 pixels from each
horizontal side. With a patch size and stride of 16, this produces a spatial
feature grid
$\mathbf{F}_t\in\mathbb{R}^{16\times12\times1280}$.

We use only the spatial image features extracted by the frozen backbone.
WiLoR's task tokens and predicted MANO and camera parameters are not
provided as inputs to our model during inference.

\subsection{Pose, Translation, and Appearance Branches}
\label{sec:supp_final_arch}

The pose, translation, and appearance branches process the backbone features
using representations suited to their respective prediction architectures.
The pose and translation branches use separate temporal convolution layers.
The pose branch retains a higher feature dimension for query-based attention,
whereas the translation branch uses a compact representation before
flattening the spatial features. The appearance branch uses only the
current-frame features.

\paragraph{Pose branch.}
For each spatial location in the backbone feature grid, the pose branch
applies temporal convolution over a three-frame window. The temporal
convolution reduces the feature dimension from 1280 to 128 while preserving
the $16\times12$ spatial structure. The resulting feature grid is represented
as 192 spatial tokens.

The pose branch uses seven task-specific queries corresponding to the root
orientation, five finger-pose groups, and MANO shape. The queries are
concatenated with the 192 spatial tokens and jointly processed by the pose
Transformer. The pose Transformer consists of two pre-normalization
Transformer blocks, each with four attention heads, a feed-forward dimension
of 512, GELU activations, and no dropout.

After the complete sequence passes through both Transformer blocks, the
final representation at each task-query position is passed to a separate
lightweight MLP decoder. The root-orientation query predicts a 3D axis-angle
rotation. Each finger query predicts the 9D axis-angle rotations of the
three joints of the corresponding finger, and the shape query predicts the
10D MANO shape parameters. The complete output consists of 48 pose
parameters and 10 shape parameters. The final layers of the query-specific
MLP decoders are initialized with zero weights and biases.

\paragraph{Translation branch.}
For each spatial location, the translation branch applies a separate
temporal convolution over the same three-frame window. It reduces the
feature dimension from 1280 to 16, producing a
$16\times12\times16$ feature grid. The feature grid is flattened into a
3072-dimensional vector and concatenated with the normalized bounding-box
center and log-scale information from the three input frames.

The concatenated feature is processed by an MLP with two 512-dimensional
hidden layers and ReLU activations. The MLP jointly predicts the
weak-perspective crop-camera parameters
$(s_{\mathrm{cam}},t_x,t_y)$, consisting of a scale and two in-plane
offsets. Following WiLoR, these parameters are analytically converted into
a 3D camera-space translation using the crop geometry and camera intrinsics,
positioning the reconstructed MANO hand in the full-image camera coordinate
system. The final output layer is initialized with zero weights and biases.

\paragraph{Appearance branch.}
For appearance prediction, we use only the current frame ($k=0$),
corresponding to a temporal window size of one. Each of the 192 spatial image
tokens is projected from 1280 to 16 dimensions. The resulting
$192\times16$ feature representation is flattened and passed through a
linear layer that jointly predicts a 16-dimensional embedding for each of
the 778 MANO vertices.

For each Gaussian, we obtain an interpolated embedding from the embeddings
of the three vertices of its associated mesh face using barycentric
interpolation. The interpolated embedding is processed by an attribute MLP
with two 256-dimensional hidden layers and ReLU activations. The MLP directly
predicts the Gaussian color coefficients and residual updates to its
canonical anisotropic log-scales.

\subsection{Mesh-Aligned Gaussian Construction}
\label{sec:supp_gaussian}

We associate each Gaussian with a triangular face of the predicted MANO
mesh. For each hand, we sample $15{,}000$ Gaussians on the mesh surface.
We first allocate Gaussians across the mesh faces and distribute the
remaining Gaussians in proportion to face area. Within each associated
triangle, the initial Gaussian position is sampled uniformly using
barycentric coordinates.

Let a Gaussian be associated with a face
$(\mathbf{v}_i,\mathbf{v}_j,\mathbf{v}_k)$, and let
$\mathbf{b}=(b_i,b_j,b_k)$ denote its initial barycentric coordinates, where
$b_i+b_j+b_k=1$. Its initial position and interpolated embedding are given by
\begin{equation}
\mathbf{p}
=
b_i\mathbf{v}_i+b_j\mathbf{v}_j+b_k\mathbf{v}_k,
\qquad
\tilde{\mathbf{e}}
=
b_i\mathbf{e}_i+b_j\mathbf{e}_j+b_k\mathbf{e}_k,
\end{equation}
where $\mathbf{e}_i,\mathbf{e}_j,\mathbf{e}_k\in\mathbb{R}^{16}$ are the
embeddings predicted at the three vertices of the associated face.

The interpolated embedding is processed by an attribute MLP with two
256-dimensional hidden layers and ReLU activations. For each Gaussian $g$,
the MLP directly predicts its zeroth-order spherical harmonic color
coefficients and a residual update $\Delta\mathbf{s}_g$ to its canonical
anisotropic log-scale. The final scale is computed as
\begin{equation}
\mathbf{s}_g
=
\exp\left(
\mathbf{s}_g^{0}
+
\tanh(\Delta\mathbf{s}_g)
\right),
\end{equation}
where $\mathbf{s}_g^{0}$ denotes the learnable canonical anisotropic
log-scale. A separate Gaussian adjustment decoder predicts residual updates
to the canonical surface attachment, as described in
\cref{sec:supp_adjustment}.

Gaussian opacity is initialized to $0.1$ and learned during offline training.
The number of Gaussians remains fixed, and we do not apply splitting,
densification, or pruning during either offline training or test-time
optimization.

\subsection{Mesh-Relative Gaussian Adjustment}
\label{sec:supp_adjustment}

The Gaussian adjustment decoder contains a single 64-dimensional hidden
layer. It predicts residual updates to the canonical surface attachment,
consisting of a bounded residual in barycentric-logit space, a displacement
along the face normal, and an in-plane rotation. These updates adjust the
position and orientation of each Gaussian while preserving its association
with the corresponding mesh face.

Let $\mathbf{q}_{b}\in\mathbb{R}^{3}$ denote the raw barycentric output of
the decoder. We obtain the bounded barycentric-logit residual as
$\Delta\mathbf{b}=0.2\tanh(\mathbf{q}_{b})$. The adjusted barycentric weights
are then computed as
\begin{equation}
\mathbf{b}'
=
\operatorname{softmax}
\left(
\log(\mathbf{b}+\epsilon)
+
\Delta\mathbf{b}
\right).
\end{equation}
The softmax ensures that the adjusted weights remain positive and sum to
one.

The decoder additionally produces two scalar outputs, $q_d$ and $q_\phi$.
We compute the normal displacement as
$\Delta d=0.002\tanh(q_d)$ and the in-plane rotation as
$\Delta\phi=(\pi/2)\tanh(q_\phi)$. These bounds limit the displacement to
$2$ mm along the face normal and the rotation to $90^\circ$ around the
normal axis.

Let $\mathbf{n}$ denote the face normal. The adjusted Gaussian position is
given by
\begin{equation}
\mathbf{p}'
=
b_i'\mathbf{v}_i
+
b_j'\mathbf{v}_j
+
b_k'\mathbf{v}_k
+
\Delta d\,\mathbf{n}.
\end{equation}
The base orientation is defined by the local frame of the associated mesh
face, with its normal axis aligned with $\mathbf{n}$. We additionally rotate
this orientation by $\Delta\phi$ around the normal axis. These bounded
adjustments allow each Gaussian to adapt to local surface and boundary
geometry while preserving its association with the hand mesh.


\subsection{Additional Loss Details}
\label{sec:supp_objectives}

\paragraph{Pose supervision weights.}
We assign unit weight to the vertex, rotation, shape, and translation
supervision terms defined in the main paper, i.e.,
$\lambda_{\mathrm{vert}}=\lambda_{\theta}=\lambda_{\beta}
=\lambda_{\tau}=1$.

\paragraph{Local Wasserstein distance.}
Following the Wasserstein Distortion formulation
\cite{qiu2024wasserstein,ozyilkan2026drop}, we compare local feature
statistics between the rendered and target hand images. Each hand region is
cropped to $128\times128$, normalized using the ImageNet statistics, and
processed by a frozen ImageNet-pretrained VGG-16 network. We use features
from \texttt{relu1\_2}, \texttt{relu2\_2}, \texttt{relu3\_3}, and
\texttt{relu4\_3}. At each level, local means and standard deviations are
computed using Gaussian-weighted neighborhoods with $\sigma=4$, scaled
according to the feature-map resolution. Their Wasserstein distances are
averaged across the four feature levels. The loss is evaluated only within
the valid hand mask, which is eroded by four pixels to reduce boundary
contamination.

\paragraph{Alpha leakage.}
We penalize rendered opacity outside the full hand silhouette obtained by
projecting the pseudo MANO mesh predicted by WiLoR. Let
$\hat{A}(\mathbf{x})$ denote the rendered alpha and let $B(\mathbf{x})$
indicate pixels outside the projected WiLoR MANO silhouette. We define
\begin{equation}
\mathcal{L}_{\mathrm{alpha}}
=
\frac{
    \sum_{\mathbf{x}}\hat{A}(\mathbf{x})B(\mathbf{x})
}{
    \sum_{\mathbf{x}}\hat{A}(\mathbf{x})+\epsilon
}.
\end{equation}
Because the loss penalizes opacity only outside the full projected hand
silhouette, hand regions occluded by interacting objects are not penalized
as long as they lie within the silhouette.

\subsection{Training and Test-Time Optimization}
\label{sec:supp_training}

\paragraph{Optimization schedule.}
We use Adam to optimize our model. We pretrain the pose and translation
branches for 300K iterations and the appearance branch for 20K iterations.
We then jointly optimize the pose, translation, and appearance branches for
20K iterations. We use a batch size of four for all offline training stages,
while keeping the pretrained vision backbone frozen.

During joint training, the base learning rate is exponentially decayed from
$5\times10^{-3}$ to $2\times10^{-3}$. We apply learning-rate multipliers of
$0.1$, $0.01$, and $0.01$ to the pose, translation, and appearance branches,
respectively.

\paragraph{Dataset sampling.}
For pose and translation pretraining and joint training, we sample H2O,
TACO, and HoloAssist with probabilities of $40\%$, $40\%$, and $20\%$,
respectively. For appearance pretraining, we sample H2O, TACO, HoloAssist,
and PALM with probabilities of $50\%$, $20\%$, $20\%$, and $10\%$,
respectively.

\paragraph{Test-time optimization.}
We perform test-time optimization using the same optimization framework as
joint training, but for only 200 steps with a batch size of one. The base
learning rate is exponentially decayed from $5\times10^{-3}$ to
$5\times10^{-4}$. We apply learning-rate multipliers of $0.1$, $0.01$, and
$1.0$ to the pose, translation, and appearance branches, respectively.

\section{Dataset and Evaluation Protocols}
\label{sec:supp_protocols}

\subsection{Datasets}
\label{sec:supp_datasets}

\paragraph{H2O.}
H2O captures egocentric hand--object interactions with
RGB-D sequences, camera poses, and 3D hand poses. We use only RGB frames from
the egocentric viewpoint. Sequences 0--7 from \texttt{subject4/h1} are used as
the target test sequences, while the remaining H2O sequences are used for
offline training.

For each target sequence, every tenth frame is reserved for evaluation, and
the remaining frames are used for per-sequence fitting or adaptation. The
baselines fit a separate model to each target sequence under this protocol.
Hand-4DGS does not use any frame from these target sequences during offline
training and uses only the non-evaluation frames during the test-time
optimization described in \cref{sec:supp_training}. All methods are evaluated
on the same held-out frames.

\paragraph{ARCTIC.}
ARCTIC captures two-handed manipulation of articulated
objects using multi-view cameras and a motion-capture system. It is not used
during any stage of our offline training and is therefore used to evaluate
cross-dataset generalization. From \texttt{s01}, we evaluate the use sequences
\texttt{capsulemachine\_use\_01}, \texttt{ketchup\_use\_01},
\texttt{laptop\_use\_01}, and \texttt{notebook\_use\_01}, and the grab
sequences \texttt{espressomachine\_grab\_01},
\texttt{mixer\_grab\_01}, \texttt{phone\_grab\_01}, and
\texttt{waffleiron\_grab\_01}. Because hands are not visible from the
egocentric camera at the beginning and end of these sequences, we exclude the
first and last 100 frames of each sequence from the frames used for
per-sequence fitting or adaptation as well as those used for evaluation. We
then apply the same per-sequence fitting or adaptation and
every-tenth-frame holdout protocol used on H2O to the remaining frames for all
methods.

\subsection{Baseline Configurations}
\label{sec:supp_baselines}

We use HandNeRF++ as a radiance-field-based hand
reconstruction baseline, GaussianHand as a generalizable
hand-avatar baseline, and EVA as an optimization-based
monocular Gaussian-avatar baseline. GaussianHand performs one-shot fitting for
100 optimization steps on each target sequence. For EVA, we replace its body
template with MANO to adapt it to hand reconstruction. We otherwise retain the
official configurations and implementations of all baselines.

The baselines use hand poses and projected MANO silhouettes obtained from
WiLoR during per-sequence fitting. For Hand-4DGS, WiLoR poses are used as
pseudo-labels during offline training but are not provided as network inputs
during inference. The projected WiLoR MANO silhouette is used only for alpha
supervision during test-time optimization. For all methods, we use SAM3 masks
to identify visible hand regions and handle hand--object occlusions during
image supervision.


\end{document}